\documentclass[conference]{IEEEtran}
\IEEEoverridecommandlockouts
\usepackage{cite}
\usepackage[colorlinks]{hyperref}
\usepackage{amsmath,amssymb,amsfonts}
\usepackage{algorithmic}
\usepackage{graphicx}
\usepackage{textcomp}
\usepackage{xcolor}
\def\BibTeX{{\rm B\kern-.05em{\sc i\kern-.025em b}\kern-.08em
    T\kern-.1667em\lower.7ex\hbox{E}\kern-.125emX}}
\begin{document}

\title{STC-Net: Electroluminescence-Based Solar Cell Crack Segmentation for Power Loss Estimation\\
}




\author{%
\vspace{-1em}
\makebox[\textwidth][c]{%
\begin{minipage}{0.98\textwidth}
\centering

\begin{minipage}[t]{0.31\textwidth}
\centering
{Shanaka Ramesh Gunasekara}\\
\textit{RMIT University}\\
Melbourne, Australia\\
0000-0002-5827-4878
\end{minipage}
\hfill
\begin{minipage}[t]{0.31\textwidth}
\centering
{Akila Eranda Devanarayana}\\
\textit{University of Jaffna}\\
Jaffna, Sri Lanka\\
0000-0002-7791-7402
\end{minipage}
\hfill
\begin{minipage}[t]{0.31\textwidth}
\centering
{Imasha Guruge}\\
\textit{Clean Energy Council}\\
Melbourne, Australia\\
0009-0006-5367-3040
\end{minipage}

\vspace{1em}

\begin{minipage}[t]{0.31\textwidth}
\centering
{Nuwantha Fernando}\\
\textit{RMIT University}\\
Melbourne, Australia\\
0000-0003-1047-2936
\end{minipage}
\hspace{0.12\textwidth}
\begin{minipage}[t]{0.31\textwidth}
\centering
{Ehsan Asadi}\\
\textit{RMIT University}\\
Melbourne, Australia\\
0000-0002-4835-2828
\end{minipage}

\end{minipage}

}
}


\maketitle

\begin{abstract}
Accurate crack assessment in electroluminescence (EL) images is important for photovoltaic (PV) reliability analysis, yet existing segmentation methods often fail to capture the thin, elongated, and structurally constrained nature of crack defects. This paper proposes a \textit{\textbf{S}olar \textbf{T}opology \textbf{C}rack \textbf{Net}work} (STC-Net) that incorporates edge priors, spectral priors, and a boundary-topology refinement module to improve crack continuity and boundary preservation. The framework further extends segmentation to power-loss estimation by deriving a crack-associated inactive-area proxy from the predicted masks. Experiments on the PVEL-S dataset show that STC-Net achieves 95.98 MIoU, 98.01 MDice, and 98.00 MAcc during training, and 72.52 MIoU and 80.16 MDice on unseen test samples. These results demonstrate that STC-Net provides accurate crack localization while offering a practical link between EL-based defect segmentation and PV degradation assessment. \href{https://github.com/ShanakaRG/STC-Net-Electroluminescence-Based-Solar-Cell-Crack-Segmentation-for-Power-Loss-Estimation.git}{GitHub}
\end{abstract}


\section{Introduction}

The rapid growth of photovoltaic (PV) installations, especially rooftop systems, has heightened the need for reliable and efficient module inspection~\cite{jahn01}. Defects in PV modules can adversely affect performance, reliability, and service life, making accurate detection crucial for predictive maintenance and sustained long-term operation. Although infrared (IR) imaging is widely used for PV inspection~\cite{jahn01}, its limited resolution reduces precision in detecting small-scale defects. Electroluminescence (EL) imaging addresses this limitation by providing high-contrast visualization of fine structural defects within PV cells.

Cracks are among the most critical defects observed in EL images because they signify structural degradation that can disrupt current flow, reduce the electrically active area, and ultimately diminish power output. Consequently, EL imaging has become an important non-destructive technique for identifying crack-related damage, such as micro-cracks, inactive regions, and other hidden defects that are difficult to detect with conventional inspection approaches.



Recent advances in deep learning have led to the development of many methods for crack detection and segmentation in high-resolution EL images. In particular, semantic segmentation has shown strong potential for pixel-level localization of defective regions under complex textures and varying background patterns. Otamendi \textit{et. al} have proposed the first end-to-end frameworks for anomaly detection and segmentation in PV modules~\cite{Otamendi01}, while subsequent works have introduced attention-based designs~\cite{jiang01} to alleviate the severe imbalance between crack pixels and background pixels. In addition, several studies have adopted generic segmentation architectures such as U-Net, PSPNet, and DeepLabv3+ for this task. However, most methods rely on general-purpose segmentation backbones and do not explicitly model the structural characteristics of cracks. In practice, solar cell cracks exhibit diverse orientations, varying widths, irregular lengths, and fragmented yet connected patterns, which distinguish them from ordinary region-based objects in natural images. Specifically, physically meaningful cracks exhibit elongated, connected, topology-preserving structures rather than blob-like regions.

To address this limitation, this work proposes a crack segmentation framework that explicitly accounts for the geometric and topological nature of crack defects in EL images. Unlike conventional attention-based region segmentation methods, the proposed model incorporates orientation-aware and topology-aware components to better preserve crack continuity and boundary structure through directional multi-branch filtering and boundary-topology refinement. Beyond segmentation, this work further aims to estimate the associated power loss from the extracted crack characteristics. While most existing EL-based studies focus on defect classification or coarse severity assessment, such outputs provide limited insight into the actual spatial extent of damage and its practical electrical impact.

The main contributions of this work are threefold. First, we propose a lightweight end-to-end framework for crack and defect analysis in photovoltaic EL images, designed to capture crack-specific structural characteristics while maintaining low computational complexity. Second,  orientation-aware and topology-guided modules are introduced to support segmentation models to enhance continuity preservation and enables accurate detection of thin, elongated, irregular, and directionally varying crack patterns. Third, we extend the framework from visual defect segmentation to performance-aware assessment by estimating power loss from the predicted defective regions, thereby linking image-based crack detection with practical photovoltaic condition evaluation.

The rest of the paper is organized as follows. Section~\ref{related_works} presents an overview of the current work, and Section~\ref{sec:method} presents details of the proposed segmentation framework and power loss estimation. Results are presented in Section~\ref{results}. Finally, the paper is concluded in Section~\ref{conclusion}.  

\section{Background}
\label{related_works}

\subsection{Solar cell defect detection using EL images}

Electroluminescence (EL) imaging has become an important non-destructive tool for PV cell inspection because it reveals hidden structural defects with high visual clarity. Early studies use EL imaging to analyze properties related to solar cell efficiency, and the release of EL datasets have since stimulated growing interest in automated defect analysis.

Initial EL-based approaches mainly formulate defect analysis as a supervised classification problem using cell-level annotations. These methods rely on handcrafted image features combined with conventional machine learning classifiers. For example, Demant \textit{et al.}~\cite{Demant01} designed manual image features and used a support vector machine to classify solar cells with micro-cracks. While such methods demonstrate the feasibility of automated defect identification, their performance is constrained by the quality of handcrafted features and limited generalization across diverse defect patterns.

With the success of deep learning in computer vision, more recent studies have adopted convolutional neural networks (CNNs) for solar cell defect recognition~\cite{deitsch01}. These approaches remove the need for manual feature design and generally improve defect classification performance. However, classification-based methods produce image-level predictions~\cite{karimi01,deitsch01} and are inadequate when multiple defects occur within a single cell or when precise localization is required.

To overcome this limitation, object detection methods have been introduced to localize multiple defects by predicting bounding boxes around defect regions~\cite{zhang01,zhao01,meng01}. Although effective for coarse localization, bounding-box representations are not well suited to the irregular and fine-grained geometry of crack defects. Consequently, semantic and binary segmentation methods have received increasing attention for EL-based solar cell defect analysis~\cite{Otamendi01,jiang01,sovetkin01}. By providing pixel-level localization, these methods are better suited for accurate defect delineation. Existing segmentation studies mainly focus on crack and grid-line defects, which remain challenging because of their thin, elongated, and spatially discontinuous structures. This limitation motivates the development of segmentation models that better preserve crack continuity and capture their geometric characteristics.

\subsection{Power loss estimation due to the defects}

It is worth noting that, while a considerable body of prior work has focus on defect detection, classification, and semantic segmentation, fewer studies have address the more practically important aspect of how such defects translate into electrical performance degradation. In particular, most segmentation-oriented studies emphasize pixel-level accuracy and visual localization, but do not explicitly connect the extracted defect regions to power or energy loss assessment.

An EL-based performance estimation framework was reported by Hoffmann \textit{et al.} \cite{Hoffmann01}, who proposed a deep-learning pipeline to predict module power at the maximum power point directly from a single EL image. Their method estimated the retained relative power, denoted by $p_{\mathrm{rel}}$, and then converted it to module power according to the eq.~\ref{eq01}
\begin{equation}
P_{\mathrm{mpp}} = p_{\mathrm{rel}} P_{\mathrm{nom}},
\label{eq01}
\end{equation}
where $P_{\mathrm{nom}}$ is the nominal power of the module. Using cross-validation, they showed that a ResNet18-based regressor achieves the best performance among the tested models, outperforming support vector regression baselines. However, this framework operates at the module level and relies on global image information rather the crack segmentation. It has been indicated in their analysis that the inactive regions and fractures dominate the power prediction while cracks have lower influence in their dataset. An extended EL-based segmentation studied by linking pixel-level defect detection errors to energy-efficiency impact is presented in \cite{DUAN01}. Their work demonstrate that segmentation outputs can be translated into engineering-relevant quantities by converting pixel areas into physical defect areas and then estimating associated energy losses under assumed operating conditions. Although that framework was formulated to quantify the impact of false positives and false negatives in defect detection rather than the intrinsic electrical loss caused by cracks, it provides evidence that segmentation results can serve as a basis for downstream performance and maintenance assessment.

\section{Methodology}
\label{sec:method}

\begin{figure*}
    \centering
    \includegraphics[width=0.8\linewidth]{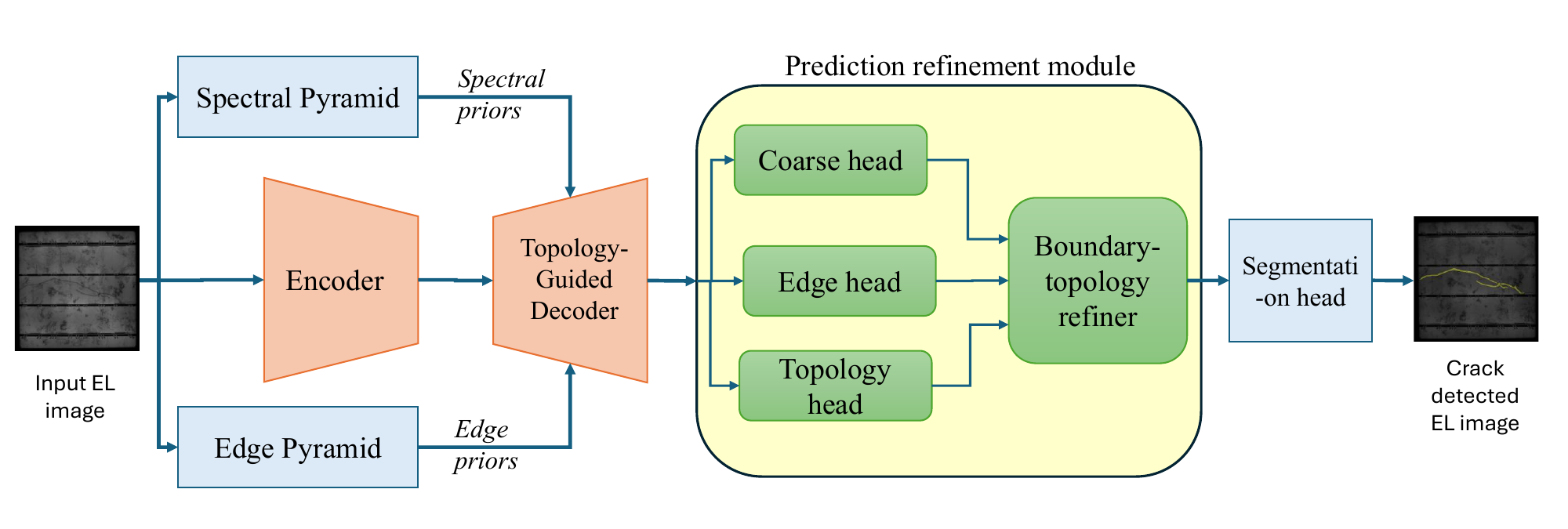}
    \caption{Overview of the proposed STC-Net framework. The input EL image is processed by the encoder, while spectral and edge pyramids extract complementary priors to guide the topology-guided decoder. The decoder output is further refined using coarse, edge, and topology heads, whose predictions are fused by the boundary-topology refiner before final prediction by the segmentation head. }
    \label{framework}
\end{figure*}
\hspace{-5em}

\subsection{Overview}
\label{subsec:overview}

The proposed \textit{\textbf{S}olar \textbf{T}opology \textbf{C}rack \textbf{Net}work}, termed as  \textbf{STC-Net}, is designed to detect cracks in EL images with higher precision. As illustrated in Fig.~\ref{framework}, the model follows an encoder-decoder architecture, while introducing two auxiliary prior extraction branches and a dedicated prediction refinement module. The encoder learns hierarchical representations from the input EL image, whereas the auxiliary branches extract complementary \emph{spectral priors} and \emph{edge priors} to guide the decoding process. The decoder then reconstructs crack-sensitive features under this multi-source guidance. Finally, three intermediate prediction heads, namely the coarse head, edge head, and topology head, are jointly exploited by a boundary-topology refiner before generating the final segmentation mask through a segmentation head. During the inference, an aditional module is introduced to predict the power loss due to the defect of region of the solar cell. add more details.


\subsection{Encoder with Auxiliary Prior Extraction}
\label{subsec:encoder}

Given an input EL image $\mathbf{X}\in\mathbb{R}^{C\times H\times W}$, where $C=1$ for grayscale images, a hierarchical U-net-like encoder is used to extract multi-scale semantic representations. The encoder is designed to capture contextual information at different resolutions while preserving crack-relevant structures, yielding the main feature representation $\mathbf{F}_{enc}$. In parallel, two auxiliary prior branches are derived directly from the input image. The first is a \emph{spectral pyramid}, which emphasizes high-frequency and residual responses to enhance faint crack patterns. The second is an \emph{edge pyramid}, which captures boundary-aware and thin structural cues. These spectral and edge priors are generated at multiple scales and injected into the decoder to guide crack localization and refinement.



\subsubsection{Edge Prior Extraction}
\label{subsec:edge_prior}

 An edge prior branch is introduced to explicitly preserve crack boundaries and thin line-like structures. Cracks in EL images often appear as weak spatially sharp discontinuities, whose local gradients remain informative even when semantic encoding suppresses fine details. Therefore, a deterministic edge extraction module is used to provide boundary-sensitive guidance to the decoder.

Given an input image $\mathbf{X}\in\mathbb{R}^{C\times H\times W}$, the edge prior is first computed using fixed Sobel filters in the horizontal and vertical directions. Let $\mathbf{K}_{x}$ and $\mathbf{K}_{y}$ denote the two $3\times3$ Sobel kernels, defined as
\begin{equation}
\mathbf{K}_{x}=
\begin{bmatrix}
1 & 0 & -1\\
2 & 0 & -2\\
1 & 0 & -1
\end{bmatrix},
\qquad
\mathbf{K}_{y}=
\begin{bmatrix}
1 & 2 & 1\\
0 & 0 & 0\\
-1 & -2 & -1
\end{bmatrix}.
\end{equation}
For each input channel, grouped convolution is applied to obtain horizontal, $\mathbf{G}_{x}=\mathbf{X} * \mathbf{K}_{x}$, and vertical $\mathbf{G}_{y}=\mathbf{X} * \mathbf{K}_{y}$ gradient responses. 
Here $*$ denotes convolution. The gradient magnitude map is then computed as in eq.~\ref{eq3}.
\begin{equation}
\mathbf{M}_{edge}=\sqrt{\mathbf{G}_{x}^{2}+\mathbf{G}_{y}^{2}+\epsilon}
\label{eq3}
\end{equation}
where $\epsilon$ is a small constant for numerical stability. This operation highlights local intensity transitions and produces an explicit boundary-enhanced representation of the crack pattern. Since the Sobel filters are fixed rather than learned, the resulting edge prior serves as a stable structural cue instead of a task-dependent feature extractor. 

However, the raw edge magnitude map is not sufficient for multi-scale decoding. Therefore, it is processed by an \emph{edge pyramid} composed of sequential convolutional stages, each performing downsampling and feature refinement. Let $\mathbf{E}^{(l)}_{edge}$ denote the output of stage $l$. The pyramid is defined eq.~\ref{eq04}.
\begin{equation}
\mathbf{E}^{(l)}_{edge} = \phi^{(l)}_{edge}\!\left(\mathbf{E}^{(l-1)}_{edge}\right),
\qquad l=1,\dots,L.
\label{eq04}
\end{equation}
where $\mathbf{E}^{(0)}_{edge}=\mathbf{M}_{edge}$ and $\phi^{(l)}_{edge}(\cdot)$ denotes the convolutional transformation at stage $l$. This produces a hierarchy of edge-aware features aligned with the decoder resolutions.

The edge prior benefits the framework in two ways. First, it preserves fine crack contours that may be blurred during repeated encoding and decoding. Second, it provides localized boundary cues that are particularly useful for thin and low-contrast defects. As a result, the decoder is guided by both semantic encoder features and explicit geometric information, leading to improved boundary precision and crack continuity.

\subsubsection{Spectral Prior Extraction}
\label{subsec:spectral_prior}

To enhance weak and fragmented crack responses, a spectral prior branch is introduced in parallel with the main encoder. Given an input image $\mathbf{X}\in\mathbb{R}^{C\times H\times W}$, a grayscale representation is first obtained as $\mathbf{X}_{gray}=\frac{1}{C}\sum_{c=1}^{C}\mathbf{X}_{c}.$

A spectral residual map is then constructed to suppress low-frequency background components and emphasize fine structural irregularities. In the Fourier formulation, the image spectrum is first computed as
$\mathbf{Z}=\mathcal{F}(\mathbf{X}_{gray})$,
and filtered by a radial high-pass mask $\mathbf{H}$ to obtain
$\mathbf{M}_{spec}=
\left|
\mathcal{F}^{-1}(\mathbf{Z}\odot\mathbf{H})
\right|.$

The resulting response is normalized and encoded through a multi-scale spectral pyramid according to the eq.~\ref{eq05},
\begin{equation}
\mathbf{E}^{(l)}_{spec}=
\phi^{(l)}_{spec}\!\left(\mathbf{E}^{(l-1)}_{spec}\right),
\qquad l=1,\dots,L.
\label{eq05}
\end{equation}
with $\mathbf{E}^{(0)}_{spec}=\widehat{\mathbf{M}}_{spec}$. These multi-scale spectral features are injected into the decoder as structural guidance. Unlike the edge prior, which mainly emphasizes boundary discontinuities, the spectral prior is more sensitive to weak, sparse, and irregular crack patterns, thereby improving segmentation robustness for low-contrast defects.

\subsection{Topology-Guided Decoder}
\label{subsec:decoder}

The decoder progressively reconstructs high-resolution crack features by combining encoder semantics with edge and spectral priors. At decoding stage $i$, let $\mathbf{D}_{i+1}$ denote the upsampled decoder feature, $\mathbf{E}_i$ the encoder skip feature, and $\mathbf{P}^{e}_i$ and $\mathbf{P}^{s}_i$ the aligned edge and spectral priors, respectively. To make them comparable, all three skip-side features are first projected into a shared channel space: $\tilde{\mathbf{E}}_i=\phi_f(\mathbf{E}_i), \quad
\tilde{\mathbf{P}}^{e}_i=\phi_e(\mathbf{P}^{e}_i), \quad
\tilde{\mathbf{P}}^{s}_i=\phi_s(\mathbf{P}^{s}_i),$
where $\phi_f,\phi_e$ and $\phi_s$ are linear projectors. A lightweight attention unit then predicts adaptive fusion weights $\alpha_f,\alpha_e,\alpha_s,$ 
\begin{equation}
[\alpha_f,\alpha_e,\alpha_s]
=
\mathrm{Softmax}\!\left(
\omega\!\left(
[\tilde{\mathbf{E}}_i,\tilde{\mathbf{P}}^{e}_i,\tilde{\mathbf{P}}^{s}_i]
\right)\right),
\end{equation}
where $\omega(\cdot)$ denotes the fusion predictor. The refined skip feature is computed as $\hat{\mathbf{E}}_i
=
\alpha_f \odot \tilde{\mathbf{E}}_i
+
\alpha_e \odot \tilde{\mathbf{P}}^{e}_i
+
\alpha_s \odot \tilde{\mathbf{P}}^{s}_i.$
It is then fused with the upsampled decoder feature to obtain $\mathbf{D}_i=\rho\!\left([\mathrm{Up}(\mathbf{D}_{i+1}),\hat{\mathbf{E}}_i]\right),$
where $\rho(\cdot)$ denotes the decoding block. Repeating this process over all decoder stages yields the final decoder representation $\mathbf{F}_{dec}=\mathbf{D}_1$
which is then passed to the prediction refinement module. This prior-guided decoding jointly exploits semantic, boundary, and high-frequency cues,  thereby improving crack continuity, boundary sharpness, and robustness to fragmented patterns.

\subsection{Prediction Refinement Module}
\label{subsec:refinement}
The final decoder feature $\mathbf{F}_{dec}$ is further refined before generating the segmentation output.  Instead of directly predicting the final mask, three complementary intermediate logits are first generated: $\mathbf{L}_{c}=h_c(\mathbf{F}_{dec}), \qquad
\mathbf{L}_{e}=h_e(\mathbf{F}_{dec}), \qquad
\mathbf{L}_{t}=h_t(\mathbf{F}_{dec}),$
where $\mathbf{L}_{c}$ represents the coarse crack-region logits, $\mathbf{L}_{e}$ represents the edge-focused logits, and $\mathbf{L}_{t}$ represents the topology-aware logits, and $h_c(\cdot)$, $h_e(\cdot)$, and $h_t(\cdot)$ denote the coarse, edge, and topology heads, respectively. These heads capture region-level crack extent, boundary evidence, and structural continuity.

The three predictions are then fused by a boundary-topology refiner: $\mathbf{F}_{ref}=r(\mathbf{L}_{c},\mathbf{L}_{e},\mathbf{L}_{t}),$
where the function $r(\cdot)$ fuses these three complementary predictions to produce the refined representation $\mathbf{F}_{ref}$. The final segmentation output is then obtained by , $\mathbf{L}_{seg}=h_{seg}(\mathbf{F}_{ref}), \qquad
\hat{\mathbf{Y}}=\sigma(\mathbf{L}_{seg}),$
where $h_{seg}(\cdot)$ denotes the final segmentation mapping, $\mathbf{L}_{seg}$ is the final segmentation logit map, $\sigma(\cdot)$ is the sigmoid function, and $\hat{\mathbf{Y}}$ is the predicted pixel-wise crack mask. This refinement strategy moves structural reasoning closer to the output space and improves the recovery of discontinuous crack segments while reducing false responses.


\subsection{Learning Objective}
\label{subsec:objective}

The network is trained under deep structural supervision. For the final prediction $\hat{\mathbf{Y}}$ and ground-truth mask $\mathbf{Y}$, the segmentation loss combines Dice and binary cross-entropy (BCE) as in eq.~\ref{eq7}:
\begin{equation}
\mathcal{L}_{seg}
=
\lambda_{1}\mathcal{L}_{Dice}(\hat{\mathbf{Y}},\mathbf{Y})
+
\lambda_{2}\mathcal{L}_{BCE}(\hat{\mathbf{Y}},\mathbf{Y}).
\label{eq7}
\end{equation}
The Dice term improves overlap under severe foreground-background imbalance, while BCE strengthens pixel-wise discrimination. To encourage accurate region, boundary, and topology learning, auxiliary supervision is applied to the intermediate heads as in eq.~\ref{eq8}:
\begin{equation}
\mathcal{L}_{aux}
=
\alpha \mathcal{L}_{c}
+
\beta \mathcal{L}_{e}
+
\gamma \mathcal{L}_{t},
\label{eq8}
\end{equation}
where $\mathcal{L}_{c}$, $\mathcal{L}_{e}$, and $\mathcal{L}_{t}$ denote the losses for the coarse, edge, and topology outputs, respectively. The total objective is given by eq.~\ref{eq9}
\begin{equation}
\mathcal{L}_{total}
=
\mathcal{L}_{seg}
+
\mathcal{L}_{aux}.
\label{eq9}
\end{equation}
This multi-branch supervision encourages the network to learn not only discriminative crack regions, but also accurate boundaries and topologically consistent crack structures.

\subsection{Crack-Informed Power Loss Estimation}
\label{subsec:power_loss}

To relate the segmented crack pattern to electrical degradation, we estimate power loss from the fraction of crack-associated inactive area in the EL image. Let $\hat{\mathbf{Y}}\in[0,1]^{H\times W}$ denote the predicted crack probability map. A binary crack mask $\mathbf{M}_{cr}$ is first obtained by thresholding $\mathbf{M}_{cr}(u,v)=\mathbb{I}\!\left(\hat{\mathbf{Y}}(u,v)\geq \tau\right),$
where $\tau$ is the inference threshold. The predicted mask is then resized to the original image resolution for area measurement. 

To normalize defect size, a cell mask $\mathbf{M}_{cell}$ is estimated. In the simplest case, the full image is treated as the valid cell area. Alternatively, the cell region is extracted from the EL image using Gaussian smoothing, Otsu thresholding, morphological closing and opening, and largest-connected-component selection. The cell area is calculated using eq.~\ref{eq10}.
\begin{equation}
A_{cell}=\sum_{u,v}\mathbf{M}_{cell}(u,v).
\label{eq10}
\end{equation}
This provides the reference area for all subsequent ratios. Because the segmentation model predicts cracks rather than electrically inactive regions directly, an inactive-area proxy is constructed from intensity and spatial proximity cues. Let $\mathbf{X}$ denote the grayscale EL image. A dark-region mask $\mathbf{M}_{dark}$ is first defined inside the cell using a percentile threshold
\begin{equation}
\tau_{d}=\mathrm{Percentile}\bigl(\{\mathbf{X}(u,v)\mid \mathbf{M}_{cell}(u,v)=1\},\,p\bigr),
\end{equation}
\begin{equation}
\mathbf{M}_{dark}(u,v)=
\mathbb{I}\!\left(\mathbf{X}(u,v)\leq \tau_{d}\right)\mathbf{M}_{cell}(u,v),
\end{equation}
where $p$ is the inactive-intensity percentile. The predicted crack mask is then dilated to form a crack influence zone $\mathbf{M}_{zone}=\mathrm{Dilate}(\mathbf{M}_{cr}),$
and the initial inactive mask $\widetilde{\mathbf{M}}_{inact}$ is obtained as $\widetilde{\mathbf{M}}_{inact}=\mathbf{M}_{dark}\odot \mathbf{M}_{zone}.$
This mask is further refined by morphological closing and removal of small connected components, and the crack mask itself is finally included in the inactive region as, $\mathbf{M}_{inact}=\max\!\left(\mathrm{Refine}(\widetilde{\mathbf{M}}_{inact}),\mathbf{M}_{cr}\right).$
Thus, the inactive region represents dark cell pixels that are spatially associated with the predicted crack. 

The crack area and inactive area are then computed within the cell region as $A_{cr}=\sum_{u,v}\mathbf{M}_{cr}(u,v)\mathbf{M}_{cell}(u,v)$ and $A_{inact}=\sum_{u,v}\mathbf{M}_{inact}(u,v)\mathbf{M}_{cell}(u,v)$
with corresponding percentages,
\begin{equation}
\rho_{cr}=100\frac{A_{cr}}{\max(A_{cell},1)}, 
\rho_{inact}=100\frac{A_{inact}}{\max(A_{cell},1)}.
\end{equation}
Here, $\rho_{cr}$ measures the directly segmented crack fraction, whereas $\rho_{inact}$ captures the broader crack-affected degraded region. 

Let $P_{nom}$ denote the nominal cell power, obtained either from a solar cell manufacturer's details or from an assumed value. The estimated power loss $\Delta P$ is modeled as linearly proportional to the inactive-area ratio defined in eq.~\ref{eq14}, 
\begin{equation}
\Delta P=P_{nom}\frac{\rho_{inact}}{100},
\label{eq14}
\end{equation}
and the remaining power is $P_{rem}=P_{nom}-\Delta P.$
Although simple, this formulation provides an interpretable bridge between image-based crack segmentation and approximate power degradation assessment.

\section{Results}
\label{results}

\subsection{Dataset}
 The PVEL-S~\cite{DUAN01} dataset was used for the training, and evaluattion of the work. This dataset is a refined subset of the PVEL-AD dataset~\cite{PVEL-D}, released by Hebei University of Technology and Beihang University for benchmarking abnormal defect detection in PV cells. It contains 1,200 polycrystalline silicon cell images with refined defect annotations, split into 840 training images and 360 test images. 


\subsection{Implementation Details}
\label{subsec:implementation}

All EL images were resized to $512\times512$ and normalized using mean $0.5$ and standard deviation $0.5$. \textit{STC-Net} used grayscale input, base channels $32$, encoder depths $[2,2,4,2]$, drop-path rate $0.1$, spectral cutoff $0.16$, and deep supervision. Training was performed for $120$ epochs with batch size $8$, learning rate $3\times10^{-4}$. At inference, the output mask was thresholded at $0.65$, with optional threshold selection from $\{0.35, 0.40, 0.45, 0.50, 0.55, 0.6,0.65,0.7,0.75 \}$ using IoU. The final loss used weights of $1.0$, $0.15$, $0.20$, $0.10$, $0.25$, and $0.20$ for the segmentation, edge, topology, consistency, auxiliary, and coarse terms, respectively.

As direct electrical measurements were unavailable for the test samples, the effect of defects was assessed a normalized power-loss surrogate. Specifically, a unit nominal power output (1 W) was assumed for all samples, and the relative loss was derived with respect to this common reference. Consequently, the reported values represent comparative measures of defect-induced degradation and should not be interpreted as absolute physical power-loss measurements.
 
\subsection{Cracks segmentation}
The proposed \textit{STC-Net} was evaluated through quantitative and qualitative analyses. For quantitative evaluation, five widely used segmentation metrics were adopted: Mean Intersection over Union (MIoU), Mean Dice (MDice), Mean Accuracy (MAcc), and Mean Precision (MPrecision). 







The Table~\ref{tab01} and Table~\ref{tab02} compare the performance of the proposed method with the state-of-the-art methods.

\begin{table}[htbp]
\caption{Validation Performance comparison on PVEL-S dataset}
\begin{center}
  \begin{tabular}{ l l l l l l } 
\hline 
Method & FPS &   MIou & MDice  & MAcc & MPrecision  \\  

\hline 
DeepLab v3~\cite{deeplapV3} & - &95.19 & 97.50 &  97.09 & 97.93 \\
CAAK-Net\cite{DUAN01} & 19.52& 96.07& 97.98 & 97.95 & 98.01 \\
\hline 
STC-Net & 41.38 & 95.98 & 98.01 & 98.00 & 97.97 \\

\hline
  \end{tabular}
   \end{center} 
    \label{tab01}
        
\end{table}

Table~\ref{tab01} shows that STC-Net achieves the best overall training performance on PVEL-S, with the highest MIoU (95.98), MDice (98.01), and MAcc (98.00), while running at 41.38 FPS. The most notable improvement is in MIoU, where STC-Net surpasses CAAK-Net by 0.91 percentage points, indicating better spatial agreement with the ground-truth crack regions. Table~\ref{tab02} further shows that this performance generalizes well at inference, yielding 80.16 MDice, 72.52 MIoU, 83.18 recall, and 84.23 precision on unseen test samples. Although prior methods do not report inference-stage results on PVEL-S, these findings show that STC-Net enables accurate and computationally practical crack segmentation.

\begin{table}[htbp]
\caption{Inference segmentation performance on PVEL-S dataset}
\begin{center}
  \begin{tabular}{ l l l l l } 
\hline 
Method &  MDice & MIoU& Mrecall& MPrecision \\  
\hline
STC-Net & 80.16 & 72.52 & 83.18 & 84.23 \\

\hline
  \end{tabular}
   \end{center} 
    \label{tab02}
        
\end{table}

\subsection{Power loss}

\begin{figure}
    \centering
    \includegraphics[width=0.9\linewidth]{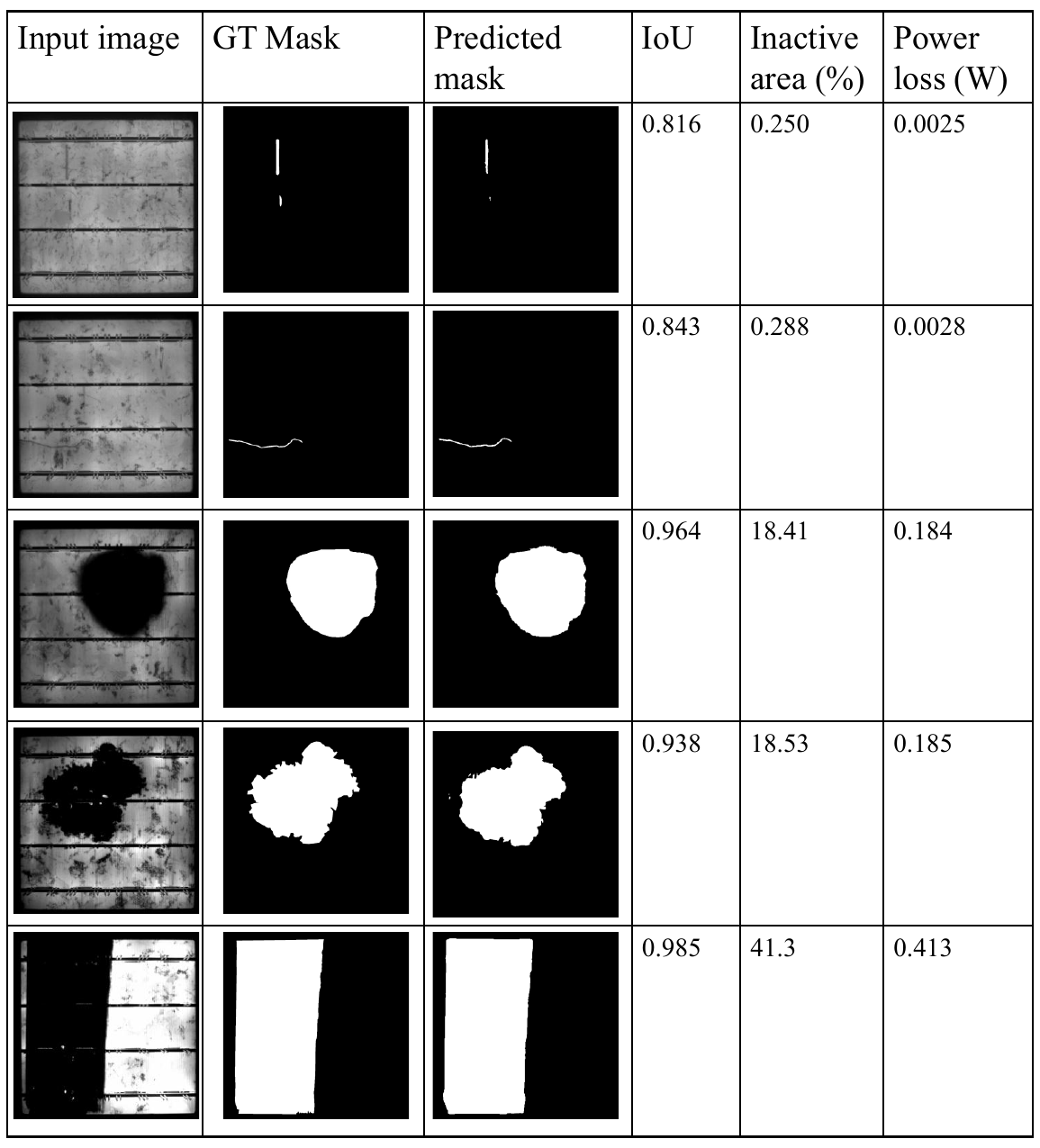}
    \caption{Qualitative STC-Net results on thin cracks and degraded regions.}
    \label{qualitative_results}
\end{figure}

Estimated degradation was analyzed at both the sample and dataset levels using representative qualitative examples. Fig.~\ref{qualitative_results} illustrates the progression from minor to severe inactive regions, while Table~\ref{tab03} presents a summary of the estimated degradation characteristics for three representative defect severity levels, namely low, medium, and high, derived from the crack-associated inactive area. 

The power-loss estimate is assumed to be linearly proportional to the defect-associated inactive-area percentage. This assumption provides a simple and interpretable first-order link between image-based segmentation and performance-oriented assessment.

The qualitative results show that STC-Net effectively preserves the thin, elongated, and structurally constrained morphology of crack defects in EL images. For fine cracks, the predicted masks closely follow the ground-truth trajectories, achieving IoU values of 0.816 and 0.843 despite the small foreground area and high sensitivity of IoU to minor boundary shifts. For larger degraded regions, STC-Net maintains strong overlap accuracy (IoU 0.938--0.985), indicating reliable recovery of defect extent and boundary structure. The inactive-area estimates further increase consistently with the predicted degradation severity, from 0.250--0.288\% with negligible power loss (0.0025--0.0028 W) to 18.41--41.3\% with higher estimated loss (0.184--0.413 W). These results support the effectiveness of the edge-, spectral-, and topology-guided design for continuity-aware crack segmentation and downstream power-loss estimation.


\begin{table}[htbp]
\caption{Estimated power loss and remaining power in each cell in different inactive area states}
\centering
\begin{tabular}{cccc}
\hline
States & Inactive area (\%) & Power loss (\%) & Remaining power (\%) \\
\hline
Low    & 3.52  & 3.5 & 96.5 \\
Medium & 40.58 & 40.6 & 59.4 \\
High   & 63.51 & 63.5 & 36.5 \\
\hline
\end{tabular}
\label{tab03}
\end{table}



\section{Conclusion}
\label{conclusion}
This paper presented STC-Net, a topology-guided cracks and defects segmentation framework for electroluminescence (EL) images of photovoltaic solar cells. Experimental results on the PVEL-S dataset showed that STC-Net achieved strong segmentation performance, reaching 95.98 MIoU, 98.01 MDice, and 98.00 MAcc during training, together with 72.52 MIoU and 80.16 MDice on unseen test samples. These results indicate that the proposed framework provides accurate and computationally practical crack localization. 


In addition, this work extends EL-based crack analysis toward power-loss estimation by introducing a crack-associated inactive-area measure. Although the model is a simplified surrogate and does not capture performance behavior, it provides an interpretable link between segmented crack regions and PV degradation. Future work will validate framework using measured electrical data, refine modeling of crack-induced inactive regions, and extend the analysis to module-level power loss, including nonlinear effects from local heating, cell interconnections, and operating conditions. By coupling crack segmentation with degradation estimation, the proposed framework bridges image-based defect analysis and engineering-oriented PV performance assessment, supporting inspection, maintenance prioritization, and reliability monitoring.

\bibliography{ref}
\bibliographystyle{IEEEtran}
\end{document}